\documentstyle[latex-acl]{article}

\title{\vspace{-0.5in}{\small In Proceedings of the Twelfth National Conference on Artificial Intelligence (AAAI-94), pp 1473, AAAI Press/The MIT Press, 1994.} \\
\LARGE\bf Building a Parser That can Afford to Interact with Semantics}

\author{Kavi Mahesh \\
 College of Computing \\
 Georgia Institute of Technology \\
 Atlanta, Georgia 30332-0280 USA \\
 mahesh@cc.gatech.edu 
}

\begin{document}

\maketitle

\noindent{}Natural language understanding programs get bogged down by the
multiplicity of possible syntactic structures while processing real
world texts that human understanders do not have much difficulty with.
In this work, I analyze the relationships between parsing strategies,
the degree of local ambiguity encountered by them, and semantic
feedback to syntax, and propose a parsing algorithm called {\em
Head-Signaled Left Corner Parsing} (HSLC) that minimizes local
ambiguities while supporting interactive syntactic and semantic
analysis.  Such a parser has been implemented in a sentence
understanding program called COMPERE (Eiselt, Mahesh, \& Holbrook
1993).

A parser could quickly eliminate many possible syntactic structures
for a sentence by using (a) the grammar to generate syntactic
expectations, (b) structural preferences such as Minimal Attachment or
Right Association, (c) feedback from semantic analysis, (d)
statistical preferences based on a corpus, or (e) case-based
preferences arising from prior texts about stereotypical situations.
None of the above strategies suffices by itself for handling real
text.

In this work, I assume that (a) we must strive to design parsing
strategies capable of analyzing general, real life text, (b) it is
beneficial to produce immediate, incremental interpretations
(`meanings') of incoming texts, and (c) semantic (and pragmatic)
analysis can provide useful feedback to syntax without requiring
unbounded resources. Given these, my objective is to design a parsing
strategy that makes the best use of linguistic preferences--both
grammatical and structural, and also semantic and conceptual
preferences, while minimizing local ambiguities. Strong cognitive
motivations for devising such a solution were presented earlier in
(Eiselt, Mahesh, \& Holbrook 1993).

The question this leads to is: When should the parser interact with
the semantic analyzer?  It should interact only when such interaction
is beneficial to one or both, that is, when one can provide some
information to the other to help reduce the number of choices being
considered.  Parsing strategies can be distinguished along a dimension
of ``eagerness'' depending on when they make commitments to a
syntactic unit and are ready for interaction with semantics. At one
end of the spectrum lies pure bottom-up parsing, being too circumspect
and precluding the use of syntactic expectations. Pure top-down
parsing, at the other end, is too eager and leads to unwarranted
backtracking. Such nondeterminism is a problem for incremental
interaction with semantics.  A combination strategy called Left Corner
(LC) Parsing has been shown to be a good middle ground for using
top-down expectations as well as avoid unnecessary early commitments
(Abney \& Johnson 1991).

LC Parsing captures the best of both bottom-up and top-down parsing by
processing the leftmost constituent of a phrase bottom-up and
predicting subsequent constituents top-down from the parent
constituent proposed using the leftmost.  LC parsing however defines a
range of strategies in the spectrum depending on the arc enumeration
strategy employed---an important distinction between different LC
parsers. In Arc Eager LC (AELC) Parsing, a node in the parse tree is
linked to its parent without waiting for all its children.  Arc
Standard LC (ASLC) Parsing, on the other hand, waits for all the
children before making attachments.

In this work, I propose an intermediate point in the LC Parsing
spectrum between ASLC and AELC strategies and argue that the proposed
point, that I call Head-Signaled LC Parsing (HSLC), turns out to be
the optimal point for efficient interaction with semantics. In this
strategy, a node is linked to its parent as soon as a particular
required child of the node is analyzed, without waiting for other
children to its right. This required unit is predefined syntactically
for each phrase; it is not the same as the standard `semantic head'.
(E.g., N is the required unit for NP, V for VP, and NP for PP.) HSLC
makes the parser wait for essential units before interacting with
semantics but does not wait for optional adjuncts (such as PP adjuncts
to NPs or VPs).

In conclusion, while LC Parsing affords incremental parsing and
optimizes memory requirements, pure LC parsing does not generate
syntactic units in an order suitable for incremental semantic
processing. HSLC, being a hybrid of LC and head-driven parsing
strategies yields the right mix to enable incremental interaction with
semantics and reduce the number of interpretations explored.
Empirical evaluation of the HSLC algorithm in the COMPERE system is
currently in progress.

\vskip 0.05in
\begin{small}
\noindent{\bf References} \\
\noindent{}Abney, S. P.; Johnson, M. 1991. Memory Requirements and Local 
Ambiguities of Parsing Strategies. {\em Journal of Psycholinguistic
Research,} 20(3):233-250.

\vskip 0.05in
\noindent{}Eiselt, K. P.; Mahesh, K.; and Holbrook, J. K. 1993. Having Your 
Cake and Eating It Too: Autonomy and Interaction in a Model of
Sentence Processing. In Proceedings of AAAI-93, 380-385.

\end{small}
\end{document}